\documentclass[tablecaption=bottom,wcp]{jmlr}

\usepackage{soul, color}
\usepackage{bbm}
\usepackage{listings}
\usepackage{subcaption}
\usepackage{textcomp}
\usepackage{fixltx2e}

\makeatletter
\newcommand{\printfnsymbol}[1]{%
  \textsuperscript{\@fnsymbol{#1}}%
}
\makeatother

\jmlrproceedings{AABI 2019}{Accepted to the 2nd Symposium on Advances in Approximate Bayesian Inference (Vancouver, Canada, 2019)}

\title[\hspace{-3mm}MultiVerse: Causal Reasoning using Importance Sampling in Probabilistic Programming]{MultiVerse: Causal Reasoning using Importance Sampling in Probabilistic Programming}

\author{\Name{Yura Perov}\nametag{\thanks{Equal contribution.} \thanks{Babylon Health. \url{https://www.babylonhealth.com/}}} \Email{yura.perov@babylonhealth.com}\\
 \Name{Logan Graham}\printfnsymbol{1} \printfnsymbol{2} \thanks{At the time of writing this work, Logan is also a DPhil (PhD) student at the University of Oxford. Note that all work for this paper has been done by Logan while he has been working full-time at Babylon Health AI Research.} \Email{logan.graham@babylonhealth.com}\\
 \Name{Kostis Gourgoulias}\printfnsymbol{2} \Email{kostis.gourgoulias@babylonhealth.com}\\
 \Name{Jonathan G. Richens}\printfnsymbol{2} \Email{jonathan.richens@babylonhealth.com}\\
 \Name{{Ciar\'{a}n M.} Lee}\printfnsymbol{2} \thanks{University College London, United Kingdom.} \Email{ciaran.lee@babylonhealth.com}\\
 \Name{Adam Baker}\printfnsymbol{2} \Email{adam.baker@babylonhealth.com}\\
 \Name{Saurabh Johri}\printfnsymbol{2} \Email{saurabh.johri@babylonhealth.com}}

\begin{document}

\maketitle

\begin{abstract}
    We elaborate on using importance sampling for causal reasoning, in particular for counterfactual inference.
    We show how this can be implemented natively in probabilistic programming.
    By considering the structure of the counterfactual query, one can significantly optimise the inference process.
    We also consider design choices to enable further optimisations.
    We introduce MultiVerse, a probabilistic programming prototype engine for approximate causal reasoning.
    We provide experimental results and compare with Pyro, an existing probabilistic programming framework with some of causal reasoning tools.
\end{abstract}

\section{Introduction and Related Work}

Machine learning has renewed interest in causal tools to aid reasoning \citep{Pearl2018}.
\emph{Counterfactuals} are particularly special causal questions as they involve the full suite of causal tools: posterior\footnote{Also called observational or associational inference. This is what we think of as inference in the canonical case: given prior and observations, associations between variables imply a posterior.}
inference and interventional reasoning \citep{Pearl2000}.
Counterfactuals are probabilistic in nature and difficult to infer, but are powerful for explanation~\citep{Wachter2017, sokol2018conversational, guidotti2018local, pedreschi2019meaningful}, fairness \cite{Kusner2017, zhang2018fairness, russell2017worlds}, policy search (e.g. \citet{Buesing2019}) and are also quantities of interest on their own (e.g. \citet{Johansson2016}). This has seen counterfactuals applied to medicine~\citep{constantinou2016value,schulam2017reliable, richens2019,Oberst2019counterfactual}, advertisement and search~\citep{Bottou2013, SwaminathanJ15, li2015counterfactual, gilotte2018offline}, translation~\citep{lawrence2017counterfactual}, and reinforcement learning~\citep{foerster2018counterfactual, Forney2017, Buesing2019}.
Consequently, counterfactual inference generally requires enhanced tools and inference procedures to incorporate both observation and intervention.
Existing frameworks are not fully equipped to handle them {\em naturally}, preventing both easy interventional reasoning, as well as optimizations that emerge when considering the full counterfactual query.

\subsection{Counterfactual Inference}
\label{sec:about_counterfactual_inference}

A counterfactual query is a \emph{what-if?} question: \emph{what would have been the outcome had I changed an input?} More formally: \emph{what would have happened in the posterior representation of a world (given observations) if in that posterior world one or more things had been forced to change?}
This is different from observational inference as it (1) fixes the context of the model to a ``posterior world'' using observational inference, but (2) then \emph{intervenes} on one or more variables \emph{in that world} by forcing each of them take a value. Interventions in the ``posterior'' world can cause variables --- previously observed or otherwise --- to take new values (i.e. ``counter'' to their observed value, or their distribution).

A counterfactual inference query can be expressed as query over $P(K'\ |\ Y=e; do(D=d))$
\footnote{We use the order of operations on the right side of the query (i.e. firstly providing the evidence, and only then performing \texttt{do}), to emphasise that it is a counterfactual query rather than an observational query with an intervention. Counterfactual notation may seem contradictory, which we discuss in Section~\ref{sec:counterfactual_notation}.}
, given a probabilistic model $M = P(X, Y)$ that consists of observed variables $Y$ and latent variables $X$, evidence values $e$ and intervention values $d$ such that $E, D \subseteq X \cup Y$. Variables $K'$ are to be predicted after we intervene in the ``posterior world'' (explained below), and they correspond to variables $K \subseteq X \cup Y$ in the original world. Most often, variables $K'$ are just variables $Y'$, and so the query becomes $P(Y'\ |\ Y=e; do(D=d))$. Following \citet{Pearl2000}, we evaluate this query in three steps:

\begin{enumerate}
    \item \textbf{Abduction (observational inference)} --- perform the query $P(X \mid Y=e)$ to receive the joint posterior over the latent variables given the evidence. This defines the ``posterior world''. The result is model $M'$, which has the same structure as model $M$ but where $X$ has been replaced by the joint posterior $X \mid Y=e$. In the new model, the previously observed variables $Y$ are not observed anymore.
    \item \textbf{Intervention (action)} --- in model $M'$, we \emph{intervene} by forcing the values of variables $D$ to values $d$. This ignores incoming edges to $D$. This results in a new model $M''$. We denote this step via the $do$-operator \citep{Pearl2000}.
    \item \textbf{Prediction} --- we predict the quantities of interest $K'$ in $M''$. Any direct or indirect descendants of $D$ need to be updated prior to estimating a value of interest.
\end{enumerate}

Abduction is the hardest part of the counterfactual procedure as it requires full inference over the joint distribution of the latents.
Abduction by exact inference is possible, but is usually difficult or intractable.
Hence, approximate inference is crucial for counterfactual inference. There are several approaches for counterfactual inference, including the twin network approach (where approximate inference, e.g. loopy belief propagation, is usually used as in \citet{Balke1994}),
single-world intervention graphs \citep{Richardson2013}, matching~\citep{Li2013}, and more.
We use the standard approach to counterfactual inference as defined above, with its three steps: abduction, intervention, and prediction.

\subsection{Probabilistic Programming}

Probabilistic programming systems \citep{Gordon2014,perov2016applications, VandeMeent2018} allow a user to: (a) write and iterate over generative probabilistic models as programs easily, (b) set arbitrary evidence for observed variables, and (c) use out-of-the-box, mostly approximate, efficient inference methods to perform queries on the models. In most cases, probabilistic programming frameworks natively support only observational inference.

\subsection{Importance Sampling for Observational Inference}

Importance sampling is an approximate inference technique that calculates the posterior $P(X\ |\ Y)$ by drawing $N$ samples $\{s_i\}$ from a proposal distribution $Q$ and accumulating the prior, proposal, and likelihood probabilities into {\it weights} $\{w_i\}$. Given this information, we can compute statistics of interests. For more details, see Section~\ref{appx:more_on_imp_sampling_for_obs_inference}.

\subsection{Related Languages and Other Related Work for Counterfactual Inference}

To the best of the authors' knowledge, no major probabilistic programming engine natively supports counterfactual inference.
However, there are (at least) three related directions of work to performing probabilistic causal inference in the settings of probabilistic programming. We provide a brief overview of these directions below, and we expand on them a little more in Section~\ref{appx:related}.

First, it has been shown~\citep{Ness2019_lecture_notes_9_6,Ness2019_homework} that in probabilistic programming languages, which support the intervention operation, such as Pyro~\citep{Bingham2018} (which has the \texttt{do} operator; or as it can be implemented in Edward~\citep{Tran2018}), it is possible to write the abduction-intervention-prediction steps in a compositional fashion to perform counterfactual inference (or causal inference, using only the intervention step).

Second, an entirely new probabilistic programming language \textsc{Omega}\textsubscript{C}~\citep{Tavares2018} for performing counterfactual inference has been recently presented, focusing on carefully considered syntax and semantics.

Third, the use of causal and counterfactual reasoning has been explored in the field of {probabilistic logic programming}~\citep{baral2007using,baral2009probabilistic,vennekens2009cp} for probabilistic logic programs, e.g. by the use of the model/query ``encoding''.

The design, interface and inference approaches presented in our paper can be employed and implemented in almost any probabilistic programming system and are mostly language-independent. The presented MultiVerse engine prototype, which is built upon existing probabilistic programming ideas and implementations, employs a {\em single}, immutable model approach for causal inference. This makes counterfactual inference more efficient by changing the \emph{inference} process, rather than changing the way in which a \emph{model} is expressed or modified into other derived models.

\section{Methods and Approach}

\subsection{Importance Sampling for Counterfactual Inference}
\label{subsec:import_sampl_for_cf}

We can perform inference for the counterfactual query $P(K' \ |\ Y = e, do(D = d))$ using importance sampling by modifying the three steps of abduction, intervention, and prediction:
\begin{enumerate}
    \item Use importance sampling to obtain a representation of the posterior distribution $P(X\ |\ Y = e)$. The key idea is that the posterior distribution is approximately represented in the form of $N$ samples, $s_1, \ldots, s_N$, and their weights, $w_1, \ldots, w_N$.
    That is, the set of tuples $\{ s_i, w_i \}_{i=1}^{N}$ is a valid approximate representation of the posterior distribution $P(X\ |\ Y = e)$.
    \item Do an intervention in each sample $s_i$ by setting values of all variables $D$ to values $d$.
    \item Propagate the effect of the intervention to any variable that depends on $D$ recursively (e.g. by re-sampling descendants of $D$).
    The new set of tuples $\{s'_i, w_i\}$ represents the intervened samples from the posterior distribution.
    Finally, compute statistics of interest, e.g. some function $f$ expected value as $E_{f(K')\ |\ Y = e, do(D = d)} = \sum_{i}f(s'_i)\cdot  \frac{w_i}{\sum_{k}{w_k}}.$
\end{enumerate}

More details on implementing this algorithm for inference for probabilistic programming and its brief computational/memory cost analysis is given in Section~\ref{sec:imp_sampling_for_cf_queries_in_pp}.

\subsection{MultiVerse and Optimisations for Importance Sampling for Counterfactual Inference}
\label{sec:optimisations}

A key contribution of this paper is MultiVerse: a probabilistic programming system for approximate counterfactual inference that exploits several speed and memory optimisations as a result of considering the counterfactual query and inference scheme.

We have designed MultiVerse to be a fully ``native'' probabilistic programming engine for causal reasoning in the sense that you define all elements abstractly and independently of each other: a model, observations and interventions, and an inference query, e.g. a counterfactual one. MultiVerse can perform observational and interventional queries, if chosen, on the same model as well.

In our system there is no double-sampling: for counterfactual inference, we draw a sample and calculate a weight from the proposal distribution given prior and observation likelihood, then we intervene on each sample itself, and predict and estimate its values of interest. On the other hand, to the best of our understanding, in Pyro\footnote{Note that we believe that the optimisation that we describe here can be implemented in future versions of Pyro (i.e. by further supporting counterfactual inference more natively) and other probabilistic programming frameworks as well.} one might generally need to redraw samples from the posterior distribution representation and one needs to manually force values of all variables (except intervened ones) in the model to their posterior values per each sample {\em (unless using the internals of Pyro traces)}. The latter resampling from already approximated representation introduces an additional approximation error.
Also, in our implementation we save on memory and time as we don't need to define any new model or inference objects beyond the original and as we don't need to pass any models between stages of the inference.

In addition, further optimisations to counterfactual queries can be done by evaluating only those parts of the probabilistic program execution trace that must be evaluated per each step of counterfactual inference. Because MultiVerse (MV) allows a version of ``lazy'' evaluation, our tests include prototypical experiments with ``Optimised MV'' where we only evaluate the necessary parts of the trace per each step. For more info, see Section~\ref{sec:different_implementations_we_used}.

\subsection{Introducing a New Class of Observable Probabilistic Procedures}
\label{sec:introducing_new_class}

Probabilistic programming frameworks handle well \texttt{OBSERVE(variable, value)} statements for observational inference: they incorporate the likelihood of the variables given observations. However, for counterfactual inference, it is generally necessary to represent the noise variables as explicit random variables in the trace because the noise variables should be part of the joint posterior that is received after the abduction step. In MultiVerse, we introduce ``Observable'' Random Procedures that are similar to regular Random Procedures but also (a) have an explicit noise variable that is the part of the program trace, and (b) have an inverse function that proposes that variable to a specific value to ``match'' the hyperparameters of the random procedure and the observation. For more details, see Section~\ref{sec:model_design_choices_obs_erps}.

\section{Experiments}

To evaluate the performance of different versions of MultiVerse (i.e. non-optimised and optimised) in terms of speed of sampling and convergence to the true counterfactual query values, we ran counterfactual queries for 1,000 Structural Causal Models which we generated for this experiment. For comparison, we also ran the same queries in Pyro. In total, we compared four approximate inference systems: ``MultiVerse'', ``MultiVerse Optimised'', Pyro without a smart proposal (``guide''), and Pyro with a smart proposal.
As discussed in Sections~\ref{sec:introducing_new_class} and~\ref{sec:model_design_choices_obs_erps}, the smart proposal forces the inference procedure to set each ``noise'' exogenous variable to a particular value such that the predicted value for an observed endogenous variable -- which follows a Delta distribution -- is the same as the observed value.
This characteristic is important to prevent significant rejection and is a consequence of the nature of the Structural Causal Model paradigm.

Our experiments show that each system converges but MultiVerse converges in less time (in terms of the constant factor) and more efficiently (in terms of the per-sample inference efficiency) than Pyro (see Section~\ref{sec:different_implementations_we_used} for more info).
MultiVerse also has the benefit of not needing to pass or store any new model objects. As expected, Pyro with a smart proposal converges more efficiently than Pyro without one.
Additionally, optimised MultiVerse performs faster than regular MultiVerse.

See Section~\ref{sec:whole_experiments_section} for more details on experiments and figures.

\section{Conclusion}

In this paper we discuss how to perform counterfactual queries using importance sampling.
Further, we introduce MultiVerse, a probabilistic programming system for causal reasoning that optimises approximate counterfactual inference.
For future work, we aim towards an approximate causal inference engine for any counterfactual query expressed in a probabilistic program, taking advantage of the structure of counterfactual queries to optimise the process of the inference and to choose one of many approximate inference methods.
As causal queries become more used in machine learning, we believe so will flexible and optimised tools that perform these types of inference.

\newpage
\appendix

\section{More Details on Experiments}
\label{sec:whole_experiments_section}

We randomly generated 1,000 Structural Causal Models~\citep{Pearl2000} (with 15 probabilistic procedures each not counting \texttt{Delta}-distribution procedures), their corresponding Bayesian networks in the form of probabilistic programs, as well as a counterfactual query of the form $\{Y, D, K'\}$ for each model.
On a 16-core EC2 instance \texttt{m4.4xlarge}, we calculated the exact counterfactual value of interest using enumeration, and then compared four approximate systems: ``MultiVerse'' (i.e. not optimised), ``MultiVerse Optimised'', Pyro without a smart proposal (``guide''), and Pyro with a smart proposal.

In the experiments, each system converges but both versions of MultiVerse experiments produce the same number of samples in less time than Pyro; for example in the experiments for 5,000 samples, ``MultiVerse'' produces 5,000 samples, on average, 92.8\% faster\footnote{Computed as \texttt{(Pyro - MV) / Pyro}. In other words, \texttt{Pyro / MV =} 14.0 times faster.} than Pyro. Additionally, ``MultiVerse Optimised'' performs computationally (i.e. in terms of the speed of producing the same number of samples) 26.1\% faster\footnote{Computed as \texttt{(MV - MVOpt) / MV}. In other words, \texttt{MV / MVOpt =} 1.35 times faster.} than regular ``MultiVerse'' on average, when compared on generating 1,000,000 samples per run.

In terms of statistical inference convergence quality (i.e. in terms of how well a sampler approximates a statistic of interest), both ``MultiVerse'' experiments have better inference convergence as well: for example, in the experiments for 5,000 samples, the mean absolute error of predicted values, when compared to the ground truth, for ``MultiVerse Optimised''\footnote{For ``MultiVerse (not optimised)'' the mean absolute error for 5,000 samples has been just slightly lower: 0.00527.} is 0.00539, while for Pyro it is 0.00723; hence ``MultiVerse Optimised'' inference convergence performance is 25.4\% more efficient\footnote{Computed as \texttt{(Pyro - MVOpt) / Pyro}. In other words, \texttt{Pyro / MVOpt =} 1.34 times more efficient.}. See Section~\ref{sec:experiment_results} for figures and details.

\subsection{Test models and counterfactual queries}

The 1,000 Structural Causal Models and their corresponding Bayesian networks with binary variables in the form of probabilistic programs were generated similar to the \emph{randomDAG} procedure in \citet{Kalisch2012}. Each Bayesian network contains 15 blocks, where each block can be of two types:
\begin{enumerate}
    \item A ``prior'' exogenous Bernoulli variable with a constant hyperparameter $p$ $\sim$ \texttt{Uniform-} \texttt{Continuous[0.3, 0.7]} that is randomly chosen during network generation.
    \item A dependent endogenous Delta-distribution variable $j$ that has a binary functional output $g(f(\ldots), \varepsilon_j)$ and a related ``noise'' exogenous Bernoulli variable $\varepsilon_j$. Function $f(x_{pa_1}, \ldots, x_{pa_{N_j}}) = \mathbbm{1}[\sum_{k=1}^{N_j}{\boldsymbol{\theta}_{j,k} x_{pa_k}} > 0.5]$\footnote{Note that function $\mathbbm{1}(\ldots)$ is an indicator function that outputs $1$ if the predicate is true or $0$ otherwise.} depends on $N_j$ parent variables $x_{pa_1}, \ldots, x_{pa_{N_j}}$. The exogenous noise variable $\varepsilon_j$ with predefined probability $q$ flips (i.e. maps from $1$ to $0$ and vice versa) the value of $f$ if $\varepsilon_j = 1$. Probability $q$ is sampled such that $q\sim$ \texttt{UniformContinuous[0.3, 0.7]} during network generation. Vector $\boldsymbol{\theta}_{j}$ is a vector with elements $\theta_{j,k} \sim Beta(5, 5)$ sampled randomly during network generation and then unit normalised.
\end{enumerate}

\begin{figure}[h]
\centering
\includegraphics[width=0.80\linewidth]{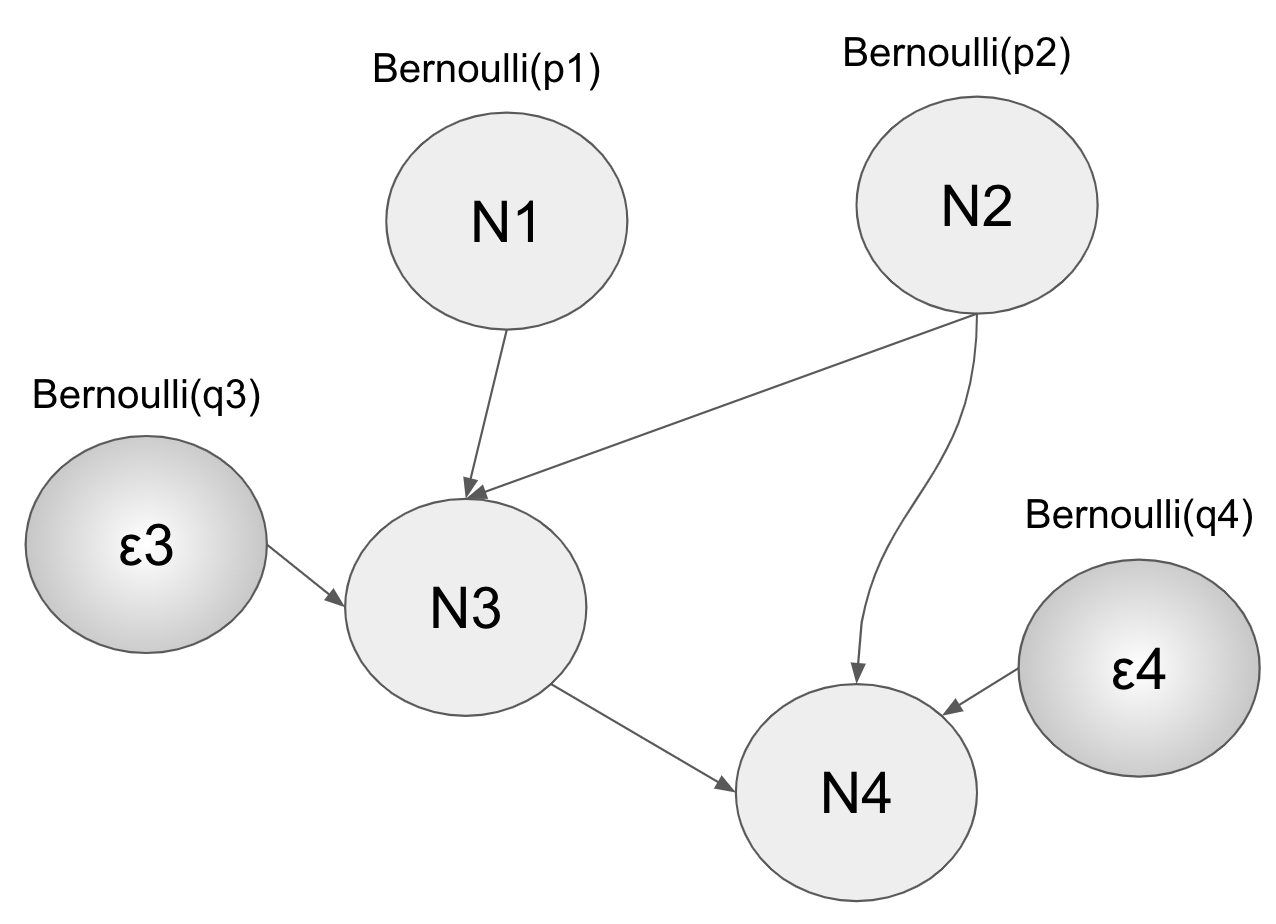}
\caption{An example of a network with similar structure to the networks which were used for the experiments. Nodes $N1$ and $N2$ are ``prior'' exogenous Bernoulli variables. Nodes $N3$ and $N4$ are ``dependent'' endogenous Delta-distribution variables, where parents of those variable include its own ``exogenous noise variable'' $\varepsilon_j$.}
\label{figure:example_exp_network}
\end{figure}

Note that every block type contains one and only one ``non-trivial'' probabilistic procedure (i.e. excluding \texttt{Delta}-distribution procedures). An example of a network with similar structure is provided in Figure~\ref{figure:example_exp_network}.

A structure of a corresponding Structural Causal Model is the same, except there is one more intermediate endogenous variable that deterministically propagates, as an identity function, the values of ``prior'' exogenous Bernoulli variables. Each counterfactual query consists of a model $G$ as defined above on which to run the query, which consists of: an evidence set $Y$ (a set of approximately 30\% of all nodes, chosen randomly) with evidence value $e$ set to a random value in $\{0, 1\}$), a single-node intervention for one node $D$ (chosen randomly) with value $d$ set to a random value in $\{0, 1\}$ or the flip of $Y_i$ if $D = Y_i$), and a node of interest $K$ (chosen randomly from all nodes except for the first two in the topological order) such that there is an active path between $K$ and $D$ with $D$ being before $K$ in the topological order.

We ran experiments with networks with only binary nodes because it simplifies the computation of the exact value of the counterfactual query result (i.e. the ground truth of it). However, it is without any loss of generality and can be extended to continuous spaces. Both Pyro and MultiVerse support both discrete and continuous variables. An example of a Gaussian model with continuous variables and code for it can be found in Section~\ref{sec:gaussian_example}.

\subsection{Different implementations that were tested}
\label{sec:different_implementations_we_used}

We run four different versions of experiments:
\begin{enumerate}
    \item ``MultiVerse'', which runs the counterfactual query as described in Section~\ref{sec:imp_sampling_for_cf_queries_in_pp}.
    
    \item ``Optimised MultiVerse'', where we calculate only variables that needs to be evaluated for abduction, intervention and prediction steps. We define lazy evaluations in our probabilistic model in Python using recursive calls. That is, we start from the variables that we must predict, observe or intervene, and evaluate only those variables and their ancestors recursively.
    
    Note that for intervened variables, we rely on MultiVerse to replace those variables with \texttt{Delta}-distributions with the intervened values, and we don't need to evaluate the parents of those intervened variables during the prediction step. An illustrative example of using MultiVerse methods for such optimisations with more details is provided in Section~\ref{sec:multiverse_optimised_code}.
    
    \item ``Pyro without guide'', in which we define a model as a probabilistic program but we don't define any guide. Because we have observations on Delta variables, this implementation leads to a lot of samples rejected.
    
    \item ``Pyro with guide'', in which we define a guide (proposal model) for ``noise'' Bernoulli variables $\{\varepsilon_i\}$. That guide forces the values of the noise variables to ensure that each observed variable is flipped or not flipped accordingly given the other parents of the observed variable and given its realised observations (i.e. to match those observations).
\end{enumerate}

\noindent Note that for Pyro we:
\begin{enumerate}
    \item Used Python Pyro package \texttt{pyro-ppl==0.4.1}.
    \item Performed two sampling steps: one for the abduction step, and another one for the intervention step where samples are drawn from the posterior. That approach of doing counterfactual inference in Pyro was also suggested in~\citep{Ness2019}, to the best of our understanding. Another, more efficient way, would be to re-use Pyro traces directly; that way we can avoid the second sampling step (e.g. by using \texttt{vectorised\_importance\_weights} which might make it significantly more efficient computationally as well). The latter approach would be then similar to the counterfactual importance sampling inference that we suggest in this paper and that is defined in Section~\ref{sec:imp_sampling_for_cf_queries_in_pp}.
    \item Used one processor, as to the best of our knowledge, parallelisation is not natively supported for importance sampling in Pyro.
    \item Pass the posterior presentation of the abducted model for intervention step as an \texttt{EmpiricalMarginal} object. In general, for very large models this might involve extra computational time/network/memory costs.
    \item For ``Pyro without guide'', if all samples for the abduction step had zero weights, we repeated that step again until at least one sample had non-zero weight.
\end{enumerate}

\subsection{Results}
\label{sec:experiment_results}

\begin{figure}[h!]
    \centering
    \includegraphics[width=0.98\linewidth]{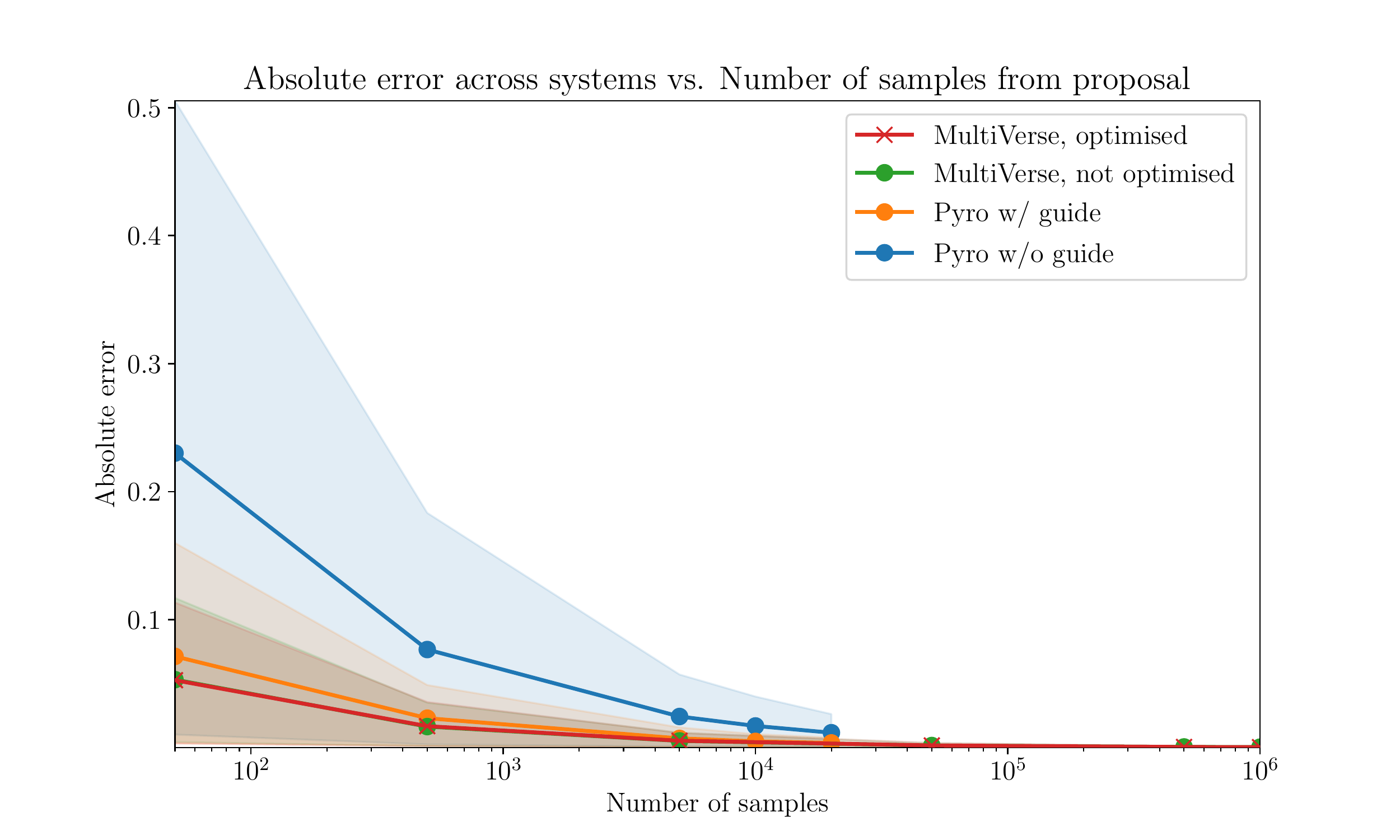}
    \caption{Convergence performance for MultiVerse and Pyro. X-axis: number of samples, Y-axis: performance that as measured by statistics computed on the absolute errors i.e. differences between predicted probabilities of the variable of the interest in the counterfactual query and the ground truth values of such queries calculated using exact enumeration. The statistics are: mean absolute errors (solid lines), as well as the 10th to 90th percentile range of absolute errors (shaded areas). The statistics are calculated across 1,000 experiments.}
    \label{fig:convergence_inference}
\end{figure}

As shown in Figure~\ref{fig:convergence_inference}, both MultiVerse (MV) and Pyro implementations seem to converge to the ground truth that has been calculated using exact enumeration. Implementations ``MV'', ``MV optimised'' have the same inference schema and hence are expected to converge similarly, inference-wise, with the same number of samples, on average. ``Pyro with guide'' is expected to converge, on average, slightly slower because of the double sampling in abduction and prediction steps as discussed in Section~\ref{sec:different_implementations_we_used}; the experimental results confirm that. Note that ``Pyro without guide'' converges, inference-wise, much slower than all three other implementations; that is expected because without a proposal (guide), a lot of samples are rejected since the observed variables don't match their observations during the abduction step.

\begin{figure}[]
    \centering
    \includegraphics[width=0.48\linewidth]{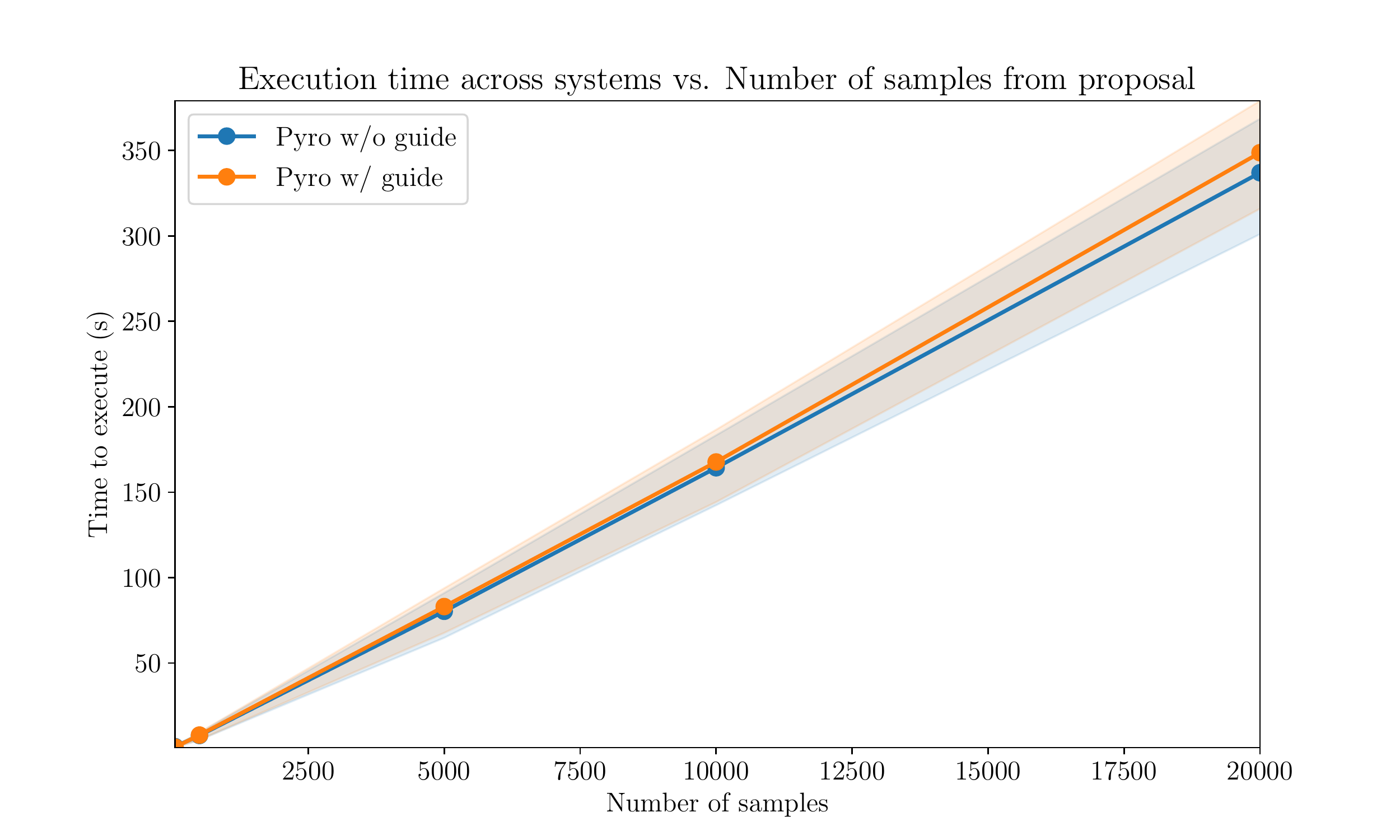}
    \includegraphics[width=0.48\linewidth]{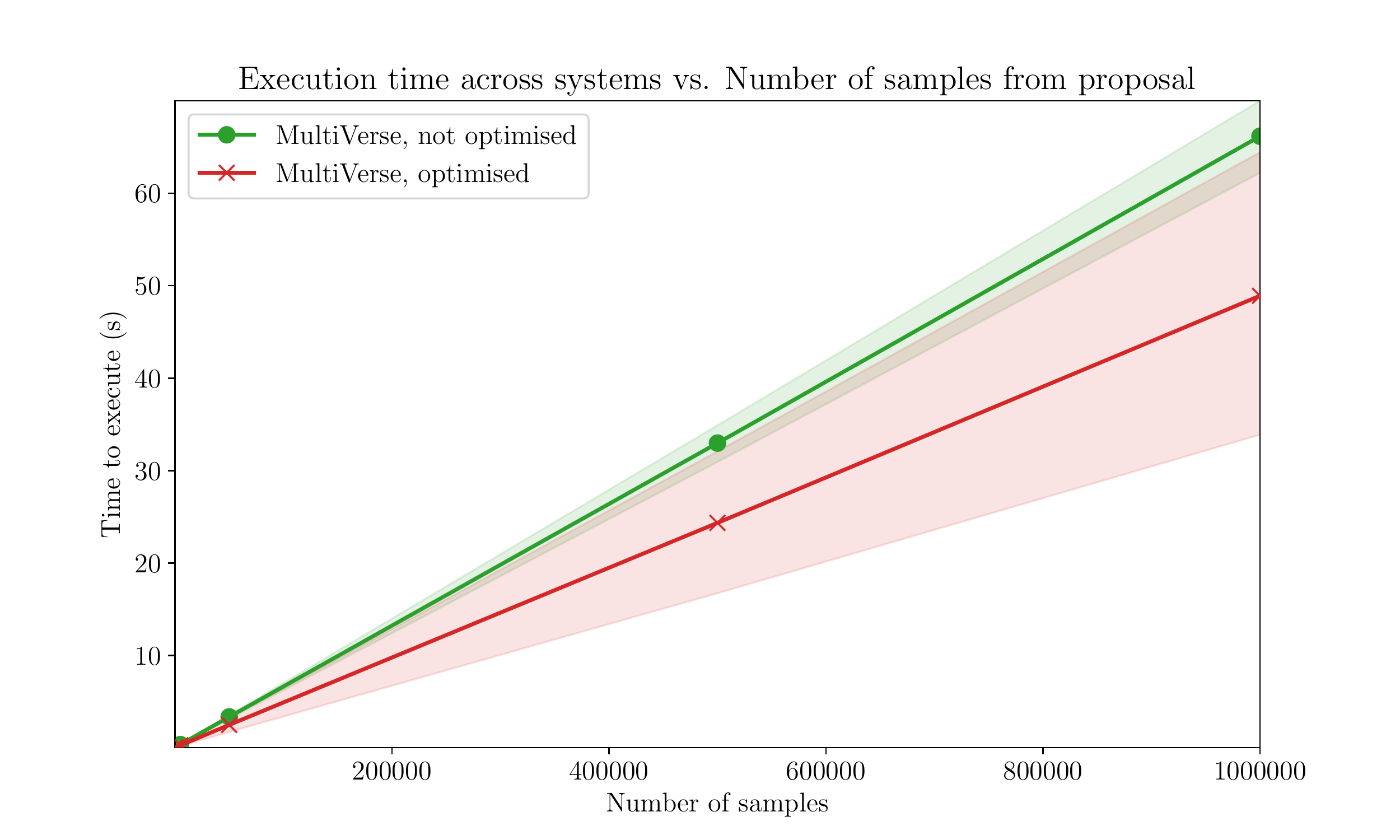}    \caption{Inference time for MultiVerse and Pyro: X-axis: number of samples, Y-axis: time in seconds.}
    \label{fig:convergence_time}
\end{figure}

Both MultiVerse implementations are significantly more efficient in terms of speed per sample than Pyro\footnote{Note that potential gains in computational efficiency might be explored in the future work by using vectorised sampling in Pyro. In our experiments, however, we aimed to use both Pyro and MultiVerse similarly to how a basic user of probabilistic programming would use them: i.e. by just writing their model as a Python program without vectorisation, at least for the first iteration over that model.} (see Figure~\ref{fig:convergence_time}). The ``Pyro with guide'' takes slightly longer (but not significantly longer) than ``Pyro without guide'', although the former is superior in terms of inference efficiency as mentioned above.

Both MultiVerse implementations support parallel importance sampling, and so both of them benefited from the fact that experiments were run on a Amazon Web Services EC2 machine with 16 cores. At the same time, as mentioned earlier, we could not find a simple way to run importance sampling in parallel in Pyro\footnote{We came to this conclusion based on the available documentation \citep{pyro_doc}. It appears there is a way to run parallel chains using MCMC but not importance sampling, to the best of our understanding. Note that someone in principal might run importance sampling samplers in parallel, but that requires additional wrappers/helpers to be implemented.}. However, if we compute average per-sample time and take into the account the number of cores (i.e. by dividing Pyro's time by 16), MultiVerse is still faster: based on experiments with 5,000 samples\footnote{As discussed earlier in the paper, for Pyro this number of samples is used twice, once for abduction step and another for prediction step.}, the average time to run 1 sample for ``Pyro w/ guide'' is 1.03833 milliseconds (already divided by 16), while for ``MultiVerse'' it is 0.07431 milliseconds and for ``MultiVerse Optimised'' it is 0.05692 milliseconds. For 1,000,000 samples, the average time to run 1 sample for ``MV'' is 0.06616 milliseconds and for ``MV'' Optimised` it is 0.04890 milliseconds.

\section{An Example Continuous Model in MultiVerse and in Pyro}
\label{sec:gaussian_example}

In this Section we provide code snippets in MultiVerse and Pyro for an example model and a counterfactual query for it.
The model is a simple Gaussian model with two latent variables $X$ and $Z$ and one emission variable $Y$ with additional Gaussian noise $\varepsilon$. It is the same as in Figure~\ref{fig:model_graph_scm}.
The counterfactual query is query $\mathbb{E}(Y'\ |\ Y = 1.2342, do(Z = -2.5236))$. That is, this counterfactual query answers the question: what would be the expected value of $Y$ if we observed $Y$ to be equal to $1.2342$ and, in that world, we have intervened $Z$ to be equal to $-2.5236$?

\newpage

\subsection{Counterfactual query model example with MultiVerse}

\begin{lstlisting}[language=Python,upquote=true]
import timeit

NUM_SAMPLES = 1000

from multiverse import (
    NormalERP,
    ObservableNormalERP,
    DeltaERP,
    observe,
    do,
    predict,
    run_inference,
)

from utils import calculate_expectation

def base_program():
    X = NormalERP(0, 1)
    Z = NormalERP(0, 1)
    Y = ObservableNormalERP(
        X.value + Z.value,
        2,
        depends_on=[X, Z],
    )
    return X, Z, Y

def program_with_data():
    X, Z, Y = base_program()

    observe(Y, 1.2342)
    do(Z, -2.5236)
    predict(Y.value, predict_counterfactual=True)

start = timeit.default_timer()
results = run_inference(program_with_data, NUM_SAMPLES)
stop = timeit.default_timer()
print('Time:', stop - start)
result = calculate_expectation(results)
print('Prediction:', result)
\end{lstlisting}

The code above in MultiVerse illustrates that MultiVerse allows the user to use ``native'' probabilistic programming in the sense that it allows him/her to ``abstractly'' and independently define a model, provide observations and interventions and perform the desired inference query (be it an observational, interventional or counterfactual query). After all of that is defined, the probabilistic programming implementation will do all inference for the user automatically. There is no need to define a new model $M'$ or store and pass a posterior $P(X \mid Y = e)$.

\subsection{Counterfactual query model example with Pyro}

This is a general way to perform counterfactual inference in Pyro using its interventional \texttt{do} operator, similar to that suggested in \citet{Ness2019}:

\begin{lstlisting}[language=Python,upquote=true]
import pyro
import torch
import numpy
import timeit

NUM_SAMPLES = 1000  # for both abduction and
                    # intervention/prediction
ROUND_DIGIT_APPR = 1  # discretisation to avoid
                      # 'infinite rejection sampling'
                      # for continuous variables
GUIDE_TO_USE = None

latent_procedure_sites = ['X', 'Z', 'Y_epsilon']

def rounder(val):
    if ROUND_DIGIT_APPR is None:
        # Don't round:
        return torch.tensor(float(val))
    else:
        return torch.tensor(
          round(float(val), ROUND_DIGIT_APPR)
        )

def extract_obs_if_any(data, var_name):
    if data is not None and var_name in data:
        return data[var_name]
    else:
        None

def model(data=None, posterior_distribution=None):
  if (
    posterior_distribution is not None
    and 'X' in posterior_distribution
  ):
    # If we are re-using a sample from a posterior,
    # we just should use the value for this variable
    # directly:
    X = posterior_distribution['X']
  else:
    X = pyro.sample(
      'X', pyro.distributions.Normal(0, 1),
      obs=extract_obs_if_any(data, 'X')
    )
  if (
    posterior_distribution is not None
    and 'Z' in posterior_distribution
  ):
    Z = posterior_distribution['Z']
  else:
    Z = pyro.sample(
      'Z', pyro.distributions.Normal(0, 1),
      obs=extract_obs_if_any(data, 'Z')
    )
  if (
    posterior_distribution is not None
    and 'Y_epsilon' in posterior_distribution
  ):
    Y_epsilon = posterior_distribution['Y_epsilon']
  else:
    Y_epsilon = pyro.sample(
      'Y_epsilon', pyro.distributions.Normal(0, 2)
    )
  discrete_Y = rounder(X + Z + Y_epsilon)

  # We must (re-)evaluate deterministic variables in any case:
  Y = pyro.sample(
    'Y',
    pyro.distributions.Delta(
        torch.tensor(discrete_Y)),
    obs=extract_obs_if_any(data, 'Y')
  )
  return X, Z, Y_epsilon, Y
  
data = {'Y': rounder(1.2342), 'X': None, 'Z': None}

start = timeit.default_timer()

# 1. Abduction
posterior = pyro.infer.Importance(
    model,
    guide=GUIDE_TO_USE,
    num_samples=NUM_SAMPLES
).run(data=data)

print('Abduction ESS:', posterior.get_ESS())

posterior = pyro.infer.EmpiricalMarginal(
  posterior,
  sites=latent_procedure_sites
)

# 2. Intervention
intervention = {'Z': -2.5236}
intervened_posterior = pyro.do(model, intervention)

# 3. Prediction
predictions_Y = []
for sample_index in range(NUM_SAMPLES):
    # We are drawing a sample from the posterior world:
    posterior_sample_vector = posterior.sample()
    # We drew that sample in a vector form;
    # now we need to transform
    # it to a dictionary of variables.
    posterior_sample = {}
    for index, var_name in enumerate(
      latent_procedure_sites
    ):
        if var_name in intervention:
            # We must ensure that we don't
            # use intervened variables
            # from its posterior:
            pass
        else:
            posterior_sample[var_name] = (
              posterior_sample_vector[index]
            )
    X, Z, Y_epsilon, Y = intervened_posterior(
      posterior_distribution=posterior_sample
    )
    predictions_Y.append(Y)

stop = timeit.default_timer()
print('Time:', stop - start)

expected_Y = numpy.mean(predictions_Y)

print('Prediction:', expected_Y)
\end{lstlisting}

Note that we have to implement the counterfactual inference using a combination of the different Pyro tools, rather than just rely on the engine and its abstractions to do it. While that is not necessarily a disadvantage, it does require a user to implement a counterfactual inference query themselves. It also requires a model to be modified and provided with the posterior values in prediction step.
Further, implementing it as described above in Pyro enforces modularity and the abduction, intervention, and prediction steps are done in isolation; they are prevented from communicating optimisations because they are individually ignorant of the full query.
Also, as mentioned in Section~\ref{sec:different_implementations_we_used}, two sampling steps are required: one for the abduction sampling step, and one for prediction step.

In addition, note that in the example above, we have to operate in a discretised continuous space for the emission variable and its observations. We achieve this with the function \texttt{rounder} and global variable \texttt{ROUND\_DIGIT\_APPR} that defines how many digits should be used when rounding the number. We have to discretise the space because otherwise it is almost impossible for the sampled value of the emission variable to match its observation.

\subsection{Using ``guide'' for more efficient counterfactual inference in Pyro}
\label{sec:Pyro_more_efficient_with_guide}

There is an alternative to the discretisation of the space. The alternative is to force the explicit noise variable $\varepsilon$ to a value that allows emission variable $Y = X + Z + \varepsilon$ to match its observation.\footnote{In this example, $\varepsilon$ is additive. In reality, it can enter in the Structural Causal Model in any form so long as there is a function \texttt{observed\_noise\_invert\_function} $f_\varepsilon$ such that $f_\varepsilon(Y, X, Z) = \varepsilon$ (it can also return none (i.e. leading to rejection) or multiple (i.e. additional sampling must be made) appropriate $\varepsilon$-s). In the case of this example, $\varepsilon = f_\varepsilon(Y, X, Z) = Y - X - Z$.} This is very similar to the idea of implemented ObserverableERPs in MultiVerse (see Section~\ref{sec:model_design_choices_obs_erps}). An example of such a guide for Pyro for this Gaussian model is provided below:

\begin{lstlisting}[language=Python,upquote=true]
def observed_normal_noise_invert_function(mean, observed_val):
  return torch.tensor(observed_val - mean)

def my_guide(data):
  X = pyro.sample('X', pyro.distributions.Normal(0, 1))
  Z = pyro.sample('Z', pyro.distributions.Normal(0, 1))

  observed_Y = data['Y']

  Y_epsilon = pyro.sample(
    'Y_epsilon',
    pyro.distributions.Delta(
      observed_normal_noise_invert_function(
        X + Z,
        observed_Y
      )
    )
  )
  discrete_Y = rounder(X + Z + Y_epsilon)

  # We must (re-)evaluate deterministic variables in any case:
  Y = pyro.sample('Y', pyro.distributions.Delta(discrete_Y))

  return X, Z, Y_epsilon, Y
\end{lstlisting}

\noindent Finally, to enable the guide (and disable the rounding over float values since with the perfect guide we have we don't need it anymore), we need to set configuration variables as follows:

\begin{lstlisting}[language=Python,upquote=true]
ROUND_DIGIT_APPR = None
GUIDE_TO_USE = my_guide
\end{lstlisting}

To compare the different options described above, we computed the effective sample size estimator~\citep{kish1965survey} of the sample weights
$$ESS = \frac{(\sum_{i=1}^{N}{w_i})^2}{\sum_{i=1}^{N}{w^2_i}}$$
for Pyro without and with a guide, as well as for MultiVerse. We ran 100 runs, each with 1,000 samples, for each option of three. The results are provided in the table below. The results illustrate that having a proposal (as a guide in Pyro) or an Observable ERP (as in MultiVerse) is crucial for efficient inference with observations and explicit noise variables.

\begin{center}
\begin{tabular}{ |c|c|c| } 
 \hline
 {\bf Experiment} & {\bf Average of ESS} & {\bf St. Dev. of ESS estimator} \\
 \hline
 Pyro without guide & 14.95 & 3.25 \\ 
 Pyro with guide & 822.61 & 9.37 \\ 
 MultiVerse & 884.73 & 4.71 \\ 
 \hline
\end{tabular}
\end{center}

\section{Details on importance sampling for counterfactual queries in probabilistic programming}

\subsection{More details on importance sampling for observational inference}
\label{appx:more_on_imp_sampling_for_obs_inference}

With importance sampling, we can approximate the observational posterior queries $P(X\ |\ Y)$ by generating $N$ samples $\{s_i\}$ from a proposal distribution $Q$ and accumulating the prior, proposal and likelihood probabilities into {\it weights} $\{w_i\}$:

$$s_i \sim \ Q_Y(X), \qquad w_i = \frac{prior(X)}{proposal(X)} \times likelihood(Y \mid X).$$

In most cases, $X$ is a vector and it can be sampled forward from the model element by element. Finally, we calculate statistics of interest using the samples and their weights. For example, we can compute the expected value of arbitrary function $f$ using the self-normalised importance sampling:
$$E_{f(X)\ |\ Y}[f(X)] = \sum_{i}f(s_i)\cdot  \frac{w_i}{\sum_{k}{w_k}}.$$

Similarly, we can do this in probabilistic programming settings, where a probabilistic program is a procedure which represents a generative model.
Each variable $x_i \in X$ and $y_i \in Y$ is represented as a random procedure. To generate a sample $s_i$ of the whole probabilistic program, we evaluate the program forward and in the process:
(a) for latent (unobserved) variables, we sample each variable value from its proposal, and incorporate the likelihood of prior and proposal into the weight;
(b) for observed variables, we incorporate the likelihood into the weight. Finally, we can compute the statistics of interest given the samples and weights.

\subsection{Note on using amortised inference}
\label{sec:amortised_inference}

Note that, as mentioned in Section~\ref{sec:about_counterfactual_inference}, the most complex step of counterfactual inference is generally the abduction step, which involves the standard joint posterior inference. The standard inference techniques for amortised approximate inference~\citep{gu2015neural,germain2015made,perov2015data,PaiWoo16,perov2016applications,le2016inference,Morris:2001:RNA:2074022.2074068,paige2016inference,ritchie2016deep,le2017using} and in particular importance sampling inference~\citep{Douetal17,walecki2018universal} can be used to facilitate the posterior inference problem.

\subsection{Importance sampling for counterfactual queries in probabilistic programming}
\label{sec:imp_sampling_for_cf_queries_in_pp}
To compute $N$ samples from counterfactual query $P(K' \ |\ Y = e, do(D = d))$ in probabilistic programming settings we need to:
\begin{enumerate}
    \setcounter{enumi}{-1}
    \item Execute the program the first time to record what variables are observed, what variables are intervened, and with what values in both cases. We also need to record what values should be predicted. The addressing scheme can be similar to the one in~\citep{lightweight_church}. Note that in this work in MultiVerse, by default, it is assumed that the structure of the program (including dependencies between variables) and the number of random choices is fixed and therefore the addresses of random variables do not change. That is because MultiVerse is based on the Python language, and in Python it is hard to track execution traces inside the program itself. Optionally, in MultiVerse, a user can also specify their own addressing scheme to account for complex control flows of the program execution and complex structure of the program.
    \item Execute that program $N$ more times without any intervention. As usual with importance sampling in probabilistic programming, we need to sample $X$ from a proposal, incorporate the prior and proposal likelihoods of $X$ into the weights, as well as the likelihoods of observed variables $Y$.
    \item Generate $N$ new samples $\{s'_i\}$ based on samples $\{s_i\}$. For each sample $s_i$, we need to re-evaluate the program but instead of sampling random variables, for each random variable we just re-use a value that was sampled for that random variable, unless that variable is in $D$ or it is any descendent variable of any variable in $D$. If the variable is in $D$, we just force the value to be the value of $d$. If the variable is an descendent (direct or indirect) of any variable in $D$, then we need to re-evaluate it (or, if it is a probabilistic procedure, to resample it).
    
    Note that in this step, weights don't need to be updated because they already represent the posterior space along with the samples. The ``counterfactual's intervention'' is intended to operate in exactly that posterior space defined by the samples and the weights\footnote{Note that if we were doing a full, exact enumeration over the posterior space (rather than importance sampling), all samples $\{s_i\}$ are expected to be unique (i.e. different from each other). However, after the intervention some samples can become identical like $s_i == s_j$ for $i \not= j$; if we want to represent the ``counterfactual'' probability space, their weights should be summed (like $w_i + w_j$) for the final representation of that probability space. For importance sampling, it does not matter too much because we usually enumerate and calculate statistics of interest over all samples $s'_i$ like $\sum_{i=1}^{N}{s'_i w_i}$, no matter whether they are unique or not.
    }\textsuperscript{,}\footnote{Note that if we are to do ``counterfactual conditioning'', which involves taking into the account additional observations on the {\em counterfactual} world after abduction and intervention, and before the prediction step, then the weights should be updated accordingly in that step as well (e.g. they should be set to 0 for the points of the counterfactual space that do not satisfy ``counterfactual conditioning'' ``observations'').
    }. (Let us remark that the ``counterfactual's intervention'' is neither an inference proposal nor an observation.)
    
    \item Predict the counterfactual variable(s) of interest. For example, for expected values it means taking the normalised weighted averages for the variable(s) of interest in the samples, as described in Clause 3 in Section~\ref{subsec:import_sampl_for_cf}.
\end{enumerate}

Based on the algorithm above, we can note that we will need $2N + 1$ evaluations of the program. Hence, the memory and computational complexity of importance sampling for counterfactual queries is the same in terms of $O$-notation as the complexity of importance sampling for ``observational'' queries as we need to evaluate each program twice. However, it takes two times more in terms of constant factor. We also can calculate $s'_i$ immediately after we calculate $s_i$ rather than firstly calculating all $\{s_i\}$ and only then calculating $\{s'_i\}$; that way the memory asymptotic complexity should be the same as for ``observational'' queries.

As with most sampling methods, we can carry out the counterfactual queries in parallel. As for the memory optimisations, instead of keeping full samples in the memory, we can discard all but predicted values for $K'$ (or, even further, we can only accumulate requested statistics of interest, e.g. a mean or variance).

We employed the optimisations mentioned in Section~\ref{sec:optimisations} by using ``lazy'' variable evaluation in ``MultiVerse Optimised'' in the experiments as discussed in Sections~\ref{sec:different_implementations_we_used} and illustrated in Section~\ref{sec:multiverse_optimised_code}.
Similar to other probabilistic programming language implementations (e.g. similar to optimisations for Church~\citep{yura2012exploiting} and Venture~\citep{mansinghka2014venture}), further/alternative optimisations can be made by tracking the dependency of a complex model graph to make sure that for computing $\{s_i\}$ and $\{s'_i\}$ we only evaluate/re-evaluate the necessary sub-parts of a program/model. That is, a more ``intelligent'' probabilistic programming engine can perform static/dynamic analysis on the computation graph (maybe even potentially a form of ``just-in-time'' compilation for inference) such that a user even don't need to make the variable evaluation ``lazy'' themselves in the model but rather the engine can determine itself what parts of the graph needs to be evaluated and when.

Also note that for efficient memory usage, ``copy-on-write'' strategies~\citep{wikipedia_copy_on_write,paige2014compilation} can be applied when making intervention/prediction on parts of the sample $s_i$ in the process of producing and using for prediction the intervened (i.e. modified) sample $s'_i$.

\subsection{Other forms of inference for counterfactual queries}

In this paper, we consider importance sampling. To be clear, counterfactual inference can be performed with any appropriate inference scheme, whether it is sampling or optimisation-based, as the abduction step can always be performed separately on its own. We focus on importance sampling for the first version of MultiVerse. Importance sampling is a conceptually straightforward\footnote{It is also an efficient way if a good proposal is used e.g. with the help of amortised inference; see Section~\ref{sec:amortised_inference} for more details.} way of showing what is perhaps one of the key insights of this paper: designing a probabilistic programming system with counterfactual and causal reasoning in mind enables further optimisations that existing general-purpose probabilistic programming system might not achieve right now (but they can be improved in the future with similar ideas). Some of these optimisations are described in Section~\ref{sec:optimisations}.

Note that using basic optimisation-based approaches instead, e.g. basic variational inference approaches, might not be efficient because one of the important considerations for counterfactual inference is the necessity to operate on the joint posterior distribution, where the joint is over all hidden variables including noise variables. Hence, approaches such as mean-field approximations are not particularly suitable, if a good precision of the final counterfactual query is desired, and more sophisticated optimisation-based approaches that preserve the relation between variables in the joint need to be used.

\subsection{Note on the contradiction in counterfactual probabilistic notation}
\label{sec:counterfactual_notation}

The notation $P(K'\ |\ Y=e; do(D=d))$ may seem contradictory at first glance, as it could be that $D \subseteq Y$ (if we intervene on a variable that we've already observed) or/and $K' \subseteq Y$ (if we are interested in a variable we've already observed).
In reality, if $D \subseteq Y$ or $K' \subseteq Y$, they are variables in model $M'$ once we have replaced the distribution $P(X)$ by $P(X \mid Y=e)$.
Hence, no contradiction occurs, but it does highlight the limited power of a standard probabilistic expression to express the intuition of the counterfactual.
This contradiction explains the name ``counterfactual'' and necessitates the three-part inference procedure.
In this short paper, we employ this ``abused'' notation to denote (left to right) first the abduction, then the intervention, in service of prediction.

\citet{Pearl2000} offers one notational resolution by denoting $K'_Y$ as the distribution of $K'$ with $X$ already updated and replaced by $P(X\mid Y=e)$. \citet{Balke1994} offers another resolution via Twin Networks, where all endogenous nodes in the Structural Causal Model are copied but share the same noise values, thus creating endogenous counterfactual counterpart variables that are separate. \citet{Richardson2013} offers Single World Intervention Graphs, a resolution similar to \citet{Balke1994} that changes the graphical representation of counterfactual and intervened variables.

\section{More Notes on Related Languages and Other Related Work for Counterfactual Inference}
\label{appx:related}

\subsection{Counterfactual Inference in Existing Probabilistic Programming
Frameworks and Comparison to MultiVerse}

Given an intervention mechanism such as exists natively in Pyro~\citep{Bingham2018} (or as can be implemented in Edward as in~\citet{Tran2018}), one can write the steps of abduction, intervention and prediction, as it has been independently shown in~\citep{Ness2019_lecture_notes_9_6,Ness2019_homework} using sampling\footnote{The suggested methodology in~\citet{Ness2019_homework} for Pyro explicitly requires resampling from the posterior to calculate counterfactual queries. For ideas on other approaches to be explored see Section~\ref{sec:different_implementations_we_used}.}.
However, the complex usage of existing methods introduces redundancy, requires model modifications, creates multiple models, and doesn't optimise inference for counterfactuals.

\subsection{Related Language: \textsc{Omega}\textsubscript{C} probabilistic programming language, syntax and semantics}

A new paper\footnote{The authors of this paper discovered the \textsc{Omega}\textsubscript{C} paper a few days before the submission of this paper.} by \citet{Tavares2018} proposes \textsc{Omega}\textsubscript{C}, a causal probabilistic programming language for inference in counterfactual generative models.
This commendable work develops its own syntax and semantics for a new language for counterfactual inference.
For future work, we are interested in:
(a) how different approximate counterfactual inference techniques operate and can be optimised in \textsc{Omega}\textsubscript{C};
(b) comparing the semantics and syntax of counterfactuals with \textsc{Omega}\textsubscript{C}, Pyro, MultiVerse and other languages, and identifying ones that are optimal for counterfactuals; and
(c) how to extend the insights from \textsc{Omega}\textsubscript{C} to other probabilistic languages and engines in order to make them more expressible and/or more efficient for counterfactuals.

\subsection{Causal and Counterfactual Reasoning in Probabilistic Logic Programming}

There is another important set of related work\footnote{We thank our reviewers for the Second Approximate Inference Symposium (see \url{http://approximateinference.org/}), to which our work has been accepted, for bringing this to our attention.} on causal reasoning and probabilistic modelling/programming, specifically in the field of {\it probabilistic logic programming}, which ``combines logic programming with probability theory as well as algorithms that operate over programs in these formalisms''~\citep{plp2019_workshop}. In particular, \cite{baral2007using} show how the probabilistic logic programming language P-log~\citep{baral2009probabilistic} can be used for causal and counterfactual reasoning e.g. with the help of special variable indexing and related encoding of a model and a query. In another paper, \cite{vennekens2009cp} develop CP-logic, a {\it logical language} for representing probabilistic causal laws in the settings of probabilistic logic programming.

\section{Model design choices for counterfactual inference in probabilistic programs}
\label{sec:model_design_choices_obs_erps}

\subsection{Observable Elementary Random Processes and their ``intelligent'' proposal distributions}

In counterfactual settings it is generally expected that all latent variables, including noise variables, should be represented explicitly. That is, the joint posterior distribution $P(X\ |\ Y = e)$ in the ``abduction'' step must account for the joint of {\em all} sources of randomness. Moreover, it is one of the requirements of structural causal models that all exogenous variables are explicitly represented.

On the other hand, the noise variables in probabilistic programming are often represented implicitly. Furthermore, often implementations of probabilistic programming systems force a user to represent them only in that implicit way. For example, an observation with the normal noise is usually absorbed into the likelihood as in the example below:

\begin{verbatim}
   Y = NORMAL(X, 1.5)
   OBSERVE(Y, 2.3)
\end{verbatim}

rather than with explicit representation of the separate, latent noise variable:

\begin{verbatim}
   EPSILON_NOISE = NORMAL(0, 1.5)
   Y = X + EPSILON_NOISE
   OBSERVE(Y, 2.3)
\end{verbatim}

There is a good reason, in general, for the implicit representation of noise variables in existing probabilistic programming frameworks (which mostly perform only observational inference) as it allows the \texttt{OBSERVE} statement to ``absorb'' the observation into the likelihood without sampling the observed variable and without intractable rejection sampling. To preserve the same benefit in our implementation but also to allow for proper counterfactual inference, we suggest to define versions of ``observable probabilistic procedures'' (OPP). For example, a Normal procedure \texttt{Normal($\mu$, $\sigma$)} can have a sibling procedure \texttt{ObservableNormal($\mu$, $\sigma$)}.

An OPP behaves in the similar way as any PP but it has two special considerations:
\begin{enumerate}
    \item An OPP must sample its noise explicitly into the program trace as an additional random variable.
    \item If an OPP is being \texttt{OBSERVE}d, then it must have a method of calculating an inverse transformation and observing that noise variable into the appropriate value, as well as calculating the marginalised likelihood of that for the trace weight.
\end{enumerate}
It is also a possible design choice to make all PPs OPPs by default, if desired.

When writing an implementation of an OPP, including its sampling and inverse methods, the similar considerations that are used for writing good proposals (i.e. ``guides'' in Pyro) can be used:
\begin{enumerate}
    \item It is mandatory to ensure that no non-zero parts of the posterior are skipped due to a proposal. For example, if there are two values of the noise variable that make the emission variable match the observation, both of those values should be sampled with non-zero probability.
    \item Note that it is okay if given some specific values of the parents of an OPP, there is no possible value that a noise variable can take to make the OPP match its observed value; in that case, that sample just should be rejected.
\end{enumerate}

\subsection{The Use of a \texttt{guide} in Pyro instead of ``Observable Elementary Random Procedures''}

For our experiments in Pyro, and generally, it is possible to use Pyro ``guide''\footnote{``Guide'' is Pyro terminology for a model that defines a proposal or variational distribution
for more efficient inference.} to force the noise variables, which must be represented explicitly, to their inversed values. An example of that is provided in Section~\ref{sec:Pyro_more_efficient_with_guide}.

\subsection{Examples of Gaussian models implying important design choices}
\label{sec:three_gaussian_examples_for_illustration}

\begin{figure}[h]
    \subfigure[]{
        \includegraphics[height=1.2in]{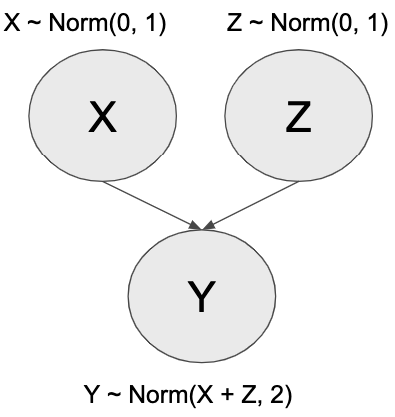}
        \label{fig:model_graph_common_pgm}
    }%
    \hfill
    \subfigure[]{
        \centering
        \includegraphics[height=1.2in]{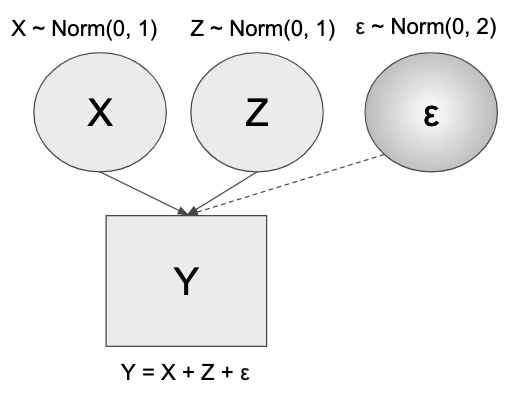}
        \label{fig:model_graph_scm}
    }%
    \hfill
    \subfigure[]{
        \centering
        \includegraphics[height=1.2in]{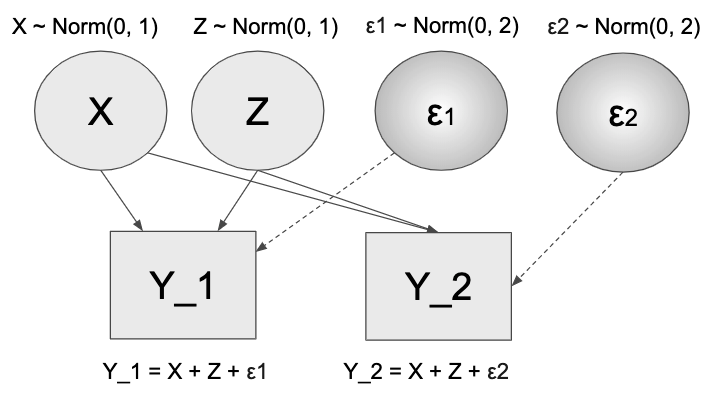}
        \label{fig:model_graph_scm_two_vars}
    }
    \hfill
    \caption{}
    \label{fig:three_models}
\end{figure}

In Figure~\ref{fig:three_models} there are three different but very similar representations/models. In all of them, there are two latent variables $X$ and $Z$, which are a priori independent and both follow prior distribution $Normal(0, 1)$. The emission variables have extra Gaussian noise $Normal(0, 2)$. Figure~\ref{fig:model_graph_common_pgm} illustrates a representation of such model with one emission variable; that is common to represent it in such way in probabilistic programming and generally for ``observational'' inference. Figure~\ref{fig:model_graph_scm} is a representation of the same model but it explicitly represents the exogenous noise variable as an independent variable (highlighted with gradient) such that the emission variable $Y$ is a {\em deterministic} function of $X$, $Z$ and $\varepsilon$ (that way, it is aligned with the general requirements of structural causal models). That way, variable $Y$ is just a sum of three variables. For the purpose of ``observational'' inference both representations have the same joint posterior $P(X, Z\ |\ Y = \hat{y})$. However, counterfactual query $P(Y'\ |\ Y = \hat{y}, do(Z = \tilde{z}))$ will be different for Figures~\ref{fig:model_graph_common_pgm} and~\ref{fig:model_graph_scm}, because in the former case the randomness over the noise has not been recorded in the joint and cannot be used for the counterfactual prediction. In other words, ``by design'', in the former case variable $Y$ has to be resampled from its prior given the posterior and intervention over its hyperparameters. Note that if anyone tries to, ``technically'', compute the ``counterfactual'' query $P(Y'\ |\ Y = \hat{y}, do(Z = \tilde{z}))$ given the model representation in Figure~\ref{fig:model_graph_common_pgm}, it will be the same as counterfactual query $P(Y_2'\ |\ Y_1 = \hat{y}, do(Z = \tilde{z}))$ in Figure~\ref{fig:model_graph_scm_two_vars} {\bf rather than} in Figure~\ref{fig:model_graph_scm} as it might had been expected. We could argue that a choice of a representation and a query should be based on an informed and mindful decision of a user (otherwise, someone accidentally would run a counterfactual query on a model as in Figure~\ref{fig:model_graph_scm_two_vars} if they implement a model as in Figure~\ref{fig:model_graph_common_pgm}, although their aim might had been to run a query on a model as in Figure~\ref{fig:model_graph_scm}). For making such a decision, someone also should consider what variables in the observed (i.e. abducted) world should still be following their prior distribution, if any at all. Figures~\ref{fig:model_graph_scm} and~\ref{fig:model_graph_scm_two_vars}, with exogenous noise separated from its deterministic child, represent quasi-Structural Causal Models (they are quasi- because the exogenous and endogenous variables for e.g. $X$ and $Y$ are combined); other model types (e.g. with not additive noise) should have different representations.

\begin{figure}[h]
\centering
\includegraphics[height=2.0in]{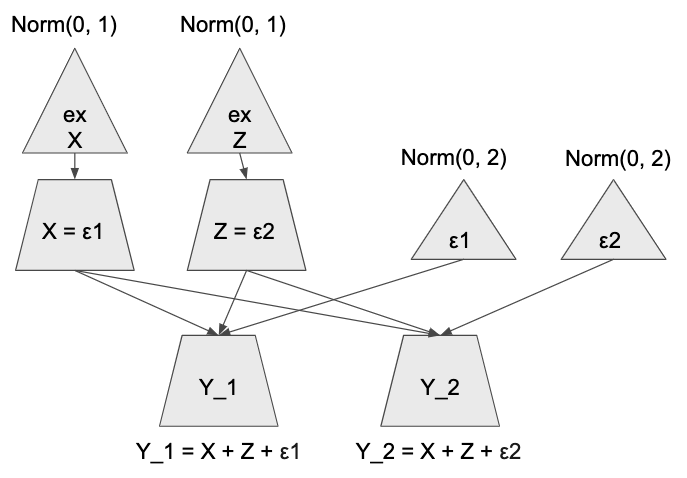}
\caption{}
\label{figure:SCM_model}
\end{figure}

For illustration, Figure~\ref{figure:SCM_model} shows a properly defined (i.e. not ``quasi-'') Structural Causal Model for Figure~\ref{fig:model_graph_scm_two_vars}. Variables $ex$, $ex$, $\varepsilon$ and $\varepsilon2$ are exogenous variables, and variables $X$, $Z$, $Y_1$ and $Y_2$ are endogenous variables. 

\subsection{Code for models in Figure~\ref{fig:three_models}}

Code for Figure~\ref{fig:model_graph_common_pgm} where, as described above, the noise for $Y$ will follow its {\bf prior} distribution:
\begin{lstlisting}[language=Python]
    X, Z = NormalERP(0, 1), NormalERP(0, 1)
    Y = NormalERP(X + Z, 2, depends_on=[X, Z])
    observe(Y, evidence_y)
    do(Z, intervened_z)
    predict(Y, counterfactual=True)
\end{lstlisting}

Code for Figure~\ref{fig:model_graph_scm} without ObservableNormalERP that leads to rejection sampling (very inefficient and generally futile (unless discretisation is used for continuous values)):
\begin{lstlisting}[language=Python]
    X, Z = NormalERP(0, 1), NormalERP(0, 1)
    EPSILON = Normal(0, 2)
    Y = X + Z + EPSILON
    # Same steps for OBSERVE, DO, PREDICT
\end{lstlisting}

Code for Figure~\ref{fig:model_graph_scm} with ObservableNormalERP, which creates an explicit Normal(0, 2) noise inside the probabilistic program execution trace and which also allows to propagate the observation inside that noise:
\begin{lstlisting}[language=Python]
    X, Z = NormalERP(0, 1), NormalERP(0, 1)
    Y = ObservableNormalERP(X + Z, 2, depends_on=[X, Z])
    # Same steps for OBSERVE, DO, PREDICT
\end{lstlisting}

Code for Figure~\ref{fig:model_graph_scm_two_vars} with ObservableNormalERP:
\begin{lstlisting}[language=Python]
    X, Z = NormalERP(0, 1), NormalERP(0, 1)
    Y1 = ObservableNormalERP(X + Z, 2, depends_on=[X, Z])
    Y2 = ObservableNormalERP(X + Z, 2, depends_on=[X, Z])
    observe(Y1, evidence_y)
    do(Z, intervened_z)
    predict(Y2, counterfactual=True)
\end{lstlisting}

\subsection{Nuance About Intervened Variables and Their Descendants in Probabilistic Programs}
\label{sec:nuance_about_intervened_variables}
Note a nuance about re-evaluating the {\em stochastic} (i.e. non-deterministic) variables that are descendants of any variables in set $D$. Following the convention suggested by Pearl et al. (e.g. see \citep{Pearl2000}) for Structural Causal Models, to the best of our understanding, it only makes sense for a \texttt{do} operation to entail an intervention on endogenous variables \citep{Pearl2000}. (It is however {\em technically} possible to intervene on exogenous variables.)

Following a similar principle in that convention, any variable that is a descendent (direct or indirect) of an intervened variable should also be an endogenous variable. That is one of the requirements of working with {\em structural causal models}. Note that in general in probabilistic programming a variable that is a descendent of any variable in $D$ can be a random variable with some hyperparameters; in other words such a variable is both an exogenous variable (defined by its own randomness (e.g. noise) that is {\bf not} expressed as a {\em separate} part of the model and hence breaks the assumptions of structural causal models) and an endogenous variable (by virtue of having hyperparameters that depend on other variables in the model).

There are at least three management strategies for this scenario:
\begin{enumerate}
    \item Be very careful when you are performing the modelling and formulate queries by ensuring you have strict structural causal models and all your queries are appropriate.
    \item Introduce checks and restrictions in a probabilistic programming language/implementation to make sure that only endogenous can be observed or intervened, as well as that any descendants of endogenous variables should be also endogenous.
    \item A ``heretical'' approach is to use this hidden randomness for the sake of using the prior conditional distributions for such variables in the counterfactual world. For example, someone can assume it might be useful if we assume that in our counterfactual query the noise is defined by its prior distribution even in the counterfactual's intervention step. However, it is only a ``hack'' due to the implementation, and if someone would like to model something like that, it might be the best to introduce proper language syntax constructions for that (e.g. by specifying what variables should be resampled from their prior (or follow a completely different distribution) in the interventional part of a counterfactual query\footnote{One way to think about that is to say that \texttt{do} operator might intervene on a variable to define a new distribution for it rather than just one value.}; or by adjusting the model and doing ``partial'' counterfactual queries as shown in one of the examples in Figure~\ref{fig:model_graph_scm_two_vars}) in Section~\ref{sec:three_gaussian_examples_for_illustration}.
\end{enumerate}

\subsection{Illustrations of ObservableNormalERP and ObservableBernoulliERP}

\begin{figure}[h!]
\centering
\includegraphics[height=2.0in]{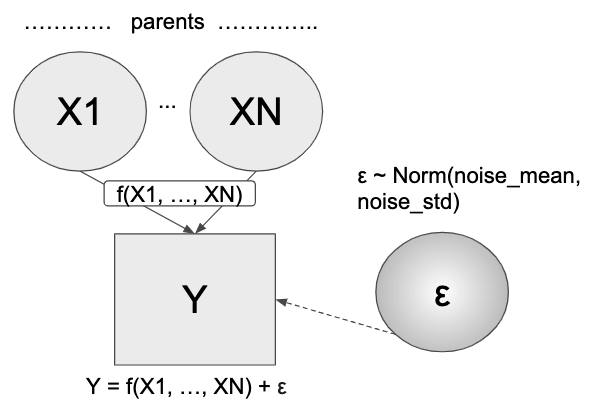}
\caption{}
\label{fig:obs_norm_erp}
\end{figure}

Figure~\ref{fig:obs_norm_erp} illustrates the Observable Normal ERP with its inverse function (for the proposal) $\varepsilon := ObservedValue - f(X1, \ldots, XN)$, if there is an observation. That way, the observation is satisfied.

\begin{figure}[h!]
\centering
\includegraphics[height=2.0in]{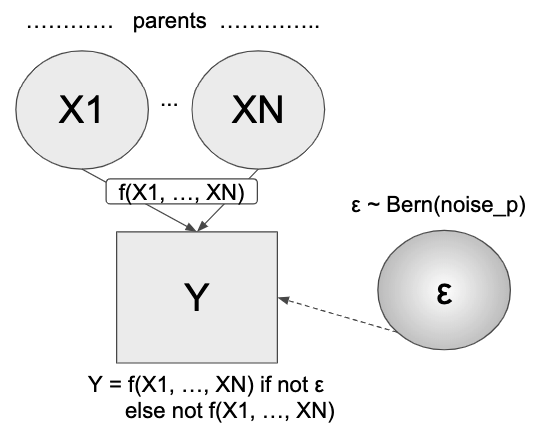}
\caption{}
\label{fig:obs_bern_erp}
\end{figure}

Figure~\ref{fig:obs_bern_erp} shows the Observable Bernoulli ERP where its noise variable $\varepsilon$ flips the output of the binary function $f(X1, \ldots, XN)$ if $\varepsilon = 1$; however, if $\varepsilon = 0$, then the emission variable $Y$ just returns the value of function $f(X1, \ldots, XN)$. The inverse function works as follows: if the output of function $f(\ldots)$ matches the observed value for $Y$, then the noise variable value is set (proposed with probability $1.0$) to value $0$ (because no flipping is required); otherwise, the noise variable value is set (proposed with probability 1.0) to value $1$ to enforce the flip. This helps to satisfy the observed value.

\subsection{More sophisticated example of an Observable ERP: the Observable Noisy OR}

A popular stochastic ``gate''/procedure is a noisy OR procedure~\citep{pearl2014probabilistic}. One of the definitions of the noisy-OR variable $Y$ given its $N$ parents $X1, \ldots, XN$ is as follows:

$$P(Y = False \mid X1, \ldots, XN) = 1.0 - \lambda_0 \prod_{i=1\ s.t.\ Xi = True}^{N}{\lambda_i}.$$

\begin{figure}[h!]
\centering
\includegraphics[height=2.0in]{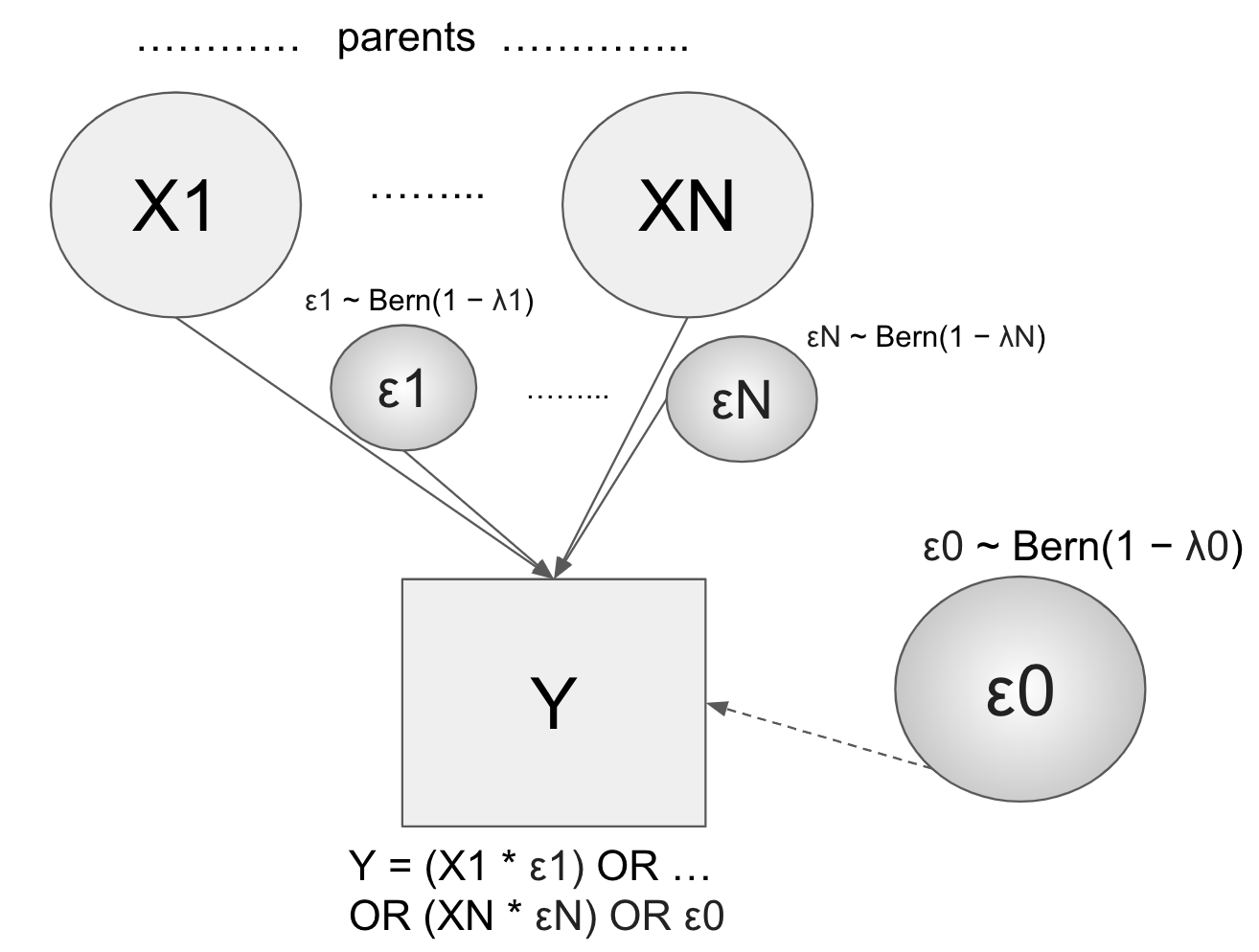}
\caption{}
\label{fig:observable_noisy_OR}
\end{figure}

Figure~\ref{fig:observable_noisy_OR} illustrates a noisy-OR gate. Each parent $j$, if active, can switch the noisy-OR variable $Y$ but with only probability $1 - \lambda_j$. In other words, there is a noisy variable associated with {\em each parent}: only if both the parent $j$ is $True$ and the associated noise variable $\varepsilon_j$ is $True$, then the noisy-OR variable $Y$ becomes $True$. There is also one more, independent, ``leak'' cause for the noisy-OR variable $Y$ to be switched on: that is if the noise (leak) variable $\varepsilon_0 = Bernoulli(1.0 - \lambda_0)$ is $True$. By that definition, the noisy-OR variable $Y$ is $False$ only if all $\varepsilon_j$, such that $j$ includes $j = 0$ and $j$ includes all parents that have state $True$, are $False$; that is exactly what the equation above calculates.

For the Observable Noisy-OR procedure, the proposal for noise variables $\varepsilon_j$ is more sophisticated. For example, a proposal might be as follows:
\begin{enumerate}
    \item If the observed value is $False$, then: (a) for any $Xi$ that is $True$, the associated noise variables $\varepsilon_j$ should be set (i.e. proposed with probability 1.0) to $False$; (b) the noise variable $\varepsilon_0$ should be set to $False$; (c) all other noise variables $\varepsilon_j$ can be sampled from any non-degenerated proposal (i.e. such that both $True$ and $False$ states have non-zero probability).
    \item If the observed value is $True$, then: (a) all $\varepsilon_j$~s.t.~$j >= 1$ can be sampled from any non-generative proposal; then (b-1) if that is enough, with the states of the parents, to enable $Y$ to be $True$, then variable $\varepsilon_0$ can be sampled from any non-generative distribution; alternatively, (b-2) if that is not enough, variable $\varepsilon_0$ must be set to $True$.
\end{enumerate}

Note that the proposal describe just above is one of many possible proposals. Another proposal might be that if the observed value is $True$, then absolutely all $\varepsilon_j$ including $j = 0$ are sampled from some non-generative proposal, and if that does not result in variable $Y$ being $True$, then that sample is just rejected (in other words, its weight will be 0).

\subsection{Interventions and possible consequences of volatile program control flows}

Note that, to the best of our understanding, for counterfactual inference, if an intervention (that happens after abduction) provokes an execution control flow change in a program, then it by default leads to the new control flow sub-trace part to be resampled from its prior. For example if in probabilistic program\footnote{To the best of our understanding, if it had been a Structural Causal Model, we might have expected variables that are used in both the \texttt{if}-consequent and \texttt{if}-alternative branches to be sampled somewhere earlier in the trace as ``exogenous'' variables like \texttt{a = Normal(1, 1)} and \texttt{b = Bernoulli(0.5)}, and then just reused inside these \texttt{if}-branches like \texttt{(if (Bernoulli 0.5) a b)}.} \texttt{(if (Bernoulli 0.1) (Normal 1 1) (Bernoulli 0.5))} the value of the predicate expression \texttt{(Bernoulli 0.1)} has been intervened from $1$ to $0$, then the value of the alternative \texttt{if}-branch expression \texttt{Bernoulli(0.5)} by default will be resampled from its prior. That resampling is similar to the resampling of the noise from its prior as discussed in Section~\ref{sec:nuance_about_intervened_variables} and might have similar implications as discussed in Section~\ref{sec:three_gaussian_examples_for_illustration}, in particular similarly to the model and query for Figure~\ref{fig:model_graph_common_pgm}.

\section{Details of MultiVerse Implementation}

MultiVerse is implemented in Python. Generally, any MultiVerse probabilistic program can be represented simple and conveniently as a Python function:

\begin{lstlisting}[language=Python]
from multiverse import (
  NormalERP,
  ObservableNormalERP,
  observe, do, predict,
  run_inference,
)

def my_model_as_probabilistic_program(param1, param2):
  X1 = NormalERP(0, 5, proposal_mean=1.5)
  X2 = NormalERP(X1 + 2, 7, depends_on=[X1])
  X3 = NormalERP(0, 8, proposal_std=2.5)
  Y1 = ObservableNormalERP(
    X1 + sin(X2) + X3, 2,
    depends_on=[X1, X2, X3])
  Y2 = ObservableNormalERP(X1 + sin(X2) + X3, 2,
    depends_on=[X1, X2, X3])
  observe(Y1, param1)
  do(X2, param2)
  predict(Y1.value)
  predict(Y2.value)
\end{lstlisting}

In the code above, we define latent variables $X1$, $X2$, $X3$. We then define emission variables $Y1$ and $Y2$, which have Gaussian noise. Because we use \verb|ObservableNormalERP|, that Gaussian noise will be represented explicitly in the trace and it will be part of the joint model posterior for further counterfactual prediction. Proposal parameters can be provided to probabilistic procedure object calls.

We then put an observation of $Y1$ for value $param1$, which is passed as an input argument to the probabilistic program. We also do an intervention for variable $X2$ to force it to value $param2$. At the end, we predict the values of $Y1$ and $Y2$.

Since in Python it is not easy to automatically track dependencies between objects (which are mutable in general), we have to explicitly specify the dependencies of each probabilistic procedure.\footnote{Note that that is required for counterfactual inference. It is not a requirement for observational inference.}\textsuperscript{,}\footnote{In the future implementations in other languages, e.g. in the subset of Clojure, the dependencies can be tracked automatically.}

Observations, similarly to other probabilistic programming languages, are provided with instruction \verb|observe(erp, value)|.\footnote{Note that currently MultiVerse allows observations and interventions only on statically defined random procedures; those procedures can't be determined based on the stochastic trace execution. It is the future work to explore that.}

By default, instruction \verb|do(erp, value, do_type=DOTYPE_CF)| performs a counterfactual intervention and instruction \verb|predict(expr, predict_counterfactual=True)| performs a counterfactual prediction.

It is also possible to perform a ``simple'' intervention on the original model by calling instruction \verb|do(erp, value, do_type=DOTYPE_IV)|, which is equivalent to modifying the original model. It is also possible to perform an ``observational'' (i.e. not counterfactual) prediction \verb|predict(expr, predict_counterfactual=False)|. Of course, it is possible to combine all of these instructions in one model/query if desired. Also note that without any counterfactual interventions \verb|do(..., do_type=DOTYPE_CF)|, the counterfactual prediction is equivalent to the ``observational'' prediction.

Performing inference is as simple as providing the evidence and the interventions, and calling \verb|run_inference| method with the selected number of samples:

\begin{lstlisting}[language=Python]
results = run_inference(
    lambda: my_model_as_probabilistic_program(3.5, 2.5),
    num_samples,
)
\end{lstlisting}

The output of \verb|run_inference| contains all predictions per each sample and samples' weights. Those can be used to compute any statistics of interest, e.g. the expected values. Method \verb|run_inference| can run inference in parallel using multiple cores.

Each ERP object creation call can be provided with its trace address by a user, e.g. \verb|X2 = NormalERP(0, 1, trace_address="X2")|. Providing such an address is optional because by default the engine uses a simple incremental addressing scheme. However if a probabilistic program has a changing control flow (e.g. there is a statement like \verb|if predicate: X2 = X1 + Normal(0, 1, ...); else: ...| such that \verb|if| statement predicate is not deterministic), then the user must use their own probabilistic procedure addressing scheme to ensure consistency for book keeping of probabilistic procedures and their values.

We chose Python for our prototype implementation because Python is a very popular language. This way, our implementation allows anyone, who knows Python, to run counterfactual queries for any probabilistic program written in Python. On the other hand, the most of the optimisations that we mentioned in Section~\ref{sec:optimisations} are harder to implement and, further, fully automatise in Python. There might be implementations of a similar engine in a restricted subset of Python or in languages like Clojure. Also, similar ideas can be implemented in existing probabilistic programming languages like Pyro, Anglican~\citep{wood2014new,tolpin2015probabilistic,tolpin2016design}, Venture, Church~\citep{goodman2012church} language engines, Gen~\citep{cusumano2018design}, and others~\citep{probabilistic_programming_website}.

\subsection{``MultiVerse Optimised'' methods to make inference faster}
\label{sec:multiverse_optimised_code}

For ``MultiVerse Optimised'' experiments as discussed in Section~\ref{sec:different_implementations_we_used}, we redefined the model such that all variables are computed in ``a lazy way'' (hence computed only if necessary), and we used some methods of MultiVerse engine that allowed us to skip computations unless they are necessary.

That is, instead of computing all variables in the model (as we did for ``MV'' and for ``Pyro'') in all steps as follows:

\begin{lstlisting}[language=Python,upquote=true]
def program():
  dict_values = {}
  for node in toposorted_nodes:
    dict_values[node] = create_prob_proc_object(node, dict_values)
    
  for var_name, val in EVIDENCE.items():
    observe(dict_values[var_name], val)
    
  for var_name, val in INTERVENTION.items():
    do(dict_values[var_name], val)
    
  predict(
    dict_values[VAR_TO_PREDICT].value,
    predict_counterfactual=True)
\end{lstlisting}

\noindent we rather compute variables only if required as shown in code snippet below. That is, if we know that we want to compute the variable of interest \texttt{VAR\_TO\_PREDICT}, so we shall compute it. Our method \texttt{compute\_var\_helper} can compute any variable, but first it will compute all its parents (and it will do so recursively for the parents of the parents, etc.). Also, we wrap calls to \texttt{compute\_var\_helper} in our method \texttt{compute\_var}, in which we rely on MultiVerse's method \texttt{compute\_procedure\_if\_necessary} to check whether we really need to compute a variable, or it has been intervened and we don't need to compute it (and hence its parents as well unless they should be computed for other reasons). MultiVerse's method \texttt{compute\_procedure\_if\_necessary(trace, procedure\_caller)} takes a variable trace and if that trace has been intervened, MultiVerse will return a \texttt{Delta}-distribution with the intervened variable's value instead; if it was not intervened, MultiVerse will compute that variable by calling function \texttt{procedure\_caller} which is provided by us. The similar logic is used for the variables that needs to be observed (i.e. \texttt{EVIDENCE} variables) or intervened (i.e. \texttt{INTERVENTION} variables). Finally, we wrap all observations in block \texttt{IF\_OBSERVE\_BLOCK} and all interventions in block \texttt{IF\_DO\_BLOCK}; that way, MultiVerse will execute those blocks only when required (e.g. we don't need to compute any observation-related parts of the program after we already did abduction step; similarly, we need to record intervention only during the initial run of the program when we record all variables that have been intervened).

\begin{lstlisting}[language=Python,upquote=true]
from multiverse import (
    IF_OBSERVE_BLOCK,
    IF_DO_BLOCK,
    compute_procedure_if_necessary,
)

def compute_var_helper(var_name, dict_values):
  for parent in cpts_parent_variables[var_name]:
    if parent not in dict_values:  # save some recursion
      compute_var(dict_values, parent)
  return create_prob_proc_object(
    var_name, dict_values)

def compute_var(dict_values, var_name):
  if var_name in dict_values:
    return dict_values[var_name]
  dict_values[var_name] = compute_procedure_if_necessary(
    var_name,
    lambda: compute_var_helper(var_name, dict_values))
  return dict_values[var_name]

def optimised_program():
  dict_values = {}
  if IF_OBSERVE_BLOCK():
    for var_name, val in EVIDENCE.items():
      observe(
        compute_var(dict_values, var_name),
        val)
  if IF_DO_BLOCK():
    for var_name, val in INTERVENTION.items():
      do(
        compute_var(dict_values, var_name),
        val)
  predict(
    compute_var(dict_values, VAR_TO_PREDICT).value,
    predict_counterfactual=True)
\end{lstlisting}

\newpage

\bibliography{main-bib}

\end{document}